\documentclass[conference]{IEEEtran}
\usepackage{times}

\usepackage[numbers]{natbib}
\usepackage{multicol}
\usepackage[bookmarks=true]{hyperref}
\usepackage{booktabs}   
\usepackage[dvipsnames]{xcolor}
\usepackage[normalem]{ulem}

\usepackage{etoolbox}
\makeatletter
\patchcmd{\@makecaption}
  {\scshape}
  {}
  {}
  {}
\makeatletter
\patchcmd{\@makecaption}
  {\\}
  {:\ }
  {}
  {}
\makeatother

\newcommand{\eq}[1]   {Eq.\,(\ref{#1})}		

\newcommand{\fig}[1]  {Fig.~\ref{#1}}		
\newcommand{\tab}[1]  {Table~\ref{#1}}		
\newcommand{\secn}[1] {Section~\ref{#1}}	

\newcommand{\pack}	{\hspace{-0.08em}}

\newcommand{\Hz}	{\ensuremath{\rm\,H\pack z}}

\newcommand{\GHz}	{\ensuremath{\rm\,G\pack H\pack z}}

\newcommand{\mm}	{\ensuremath{\rm\,m\pack m}}

\newcommand{\ms}	{\ensuremath{\rm\,m\pack s}}

\newcommand{\A}		{\ensuremath{\rm\pack A}}

\usepackage{amsmath,graphics}
\usepackage{amssymb} 

\newcommand{\cG}{\mathbf{G}}

\newcommand{\cL}{\mathcal{L}}

\newcommand{\cQ}{\mathbf{Q}}

\newcommand{\cW}{\mathbf{W}}

\newcommand{\bbR}{\mathbb{R}}

\newcommand{\fM}{M}

\newcommand{\bPhi}{\mathbf{\Phi}}

\newcommand{\bg}{\mathbf{g}}
\newcommand{\bgt}{\tilde{\mathbf{g}}}
\newcommand{\bp}{\mathbf{p}}

\newcommand{\bs}{\mathbf{s}}
\newcommand{\bT}{\mathbf{T}}

\newcommand{\by}{\mathbf{I}}

\usepackage[linesnumbered,ruled,vlined]{algorithm2e}
\usepackage{ragged2e}

\usepackage{blindtext}

\usepackage{graphicx}
\graphicspath{ {images/} }

\newcommand{\yes}{\textcolor{ForestGreen}{$\checkmark$}}
\newcommand{\no}{\textcolor{red}{$\times$}}

\begin{document}

\title{Closing the Loop for Robotic Grasping:\\A Real-time, Generative Grasp Synthesis Approach}

\author{\authorblockN{Douglas Morrison, Peter Corke and J\"urgen Leitner}
\authorblockA{Australian Centre for Robotic Vision\\
Queensland University of Technology\\
Brisbane, Australia, 4000\\
Email: douglas.morrison@hdr.qut.edu.au}
\vspace{-8mm}
}

\maketitle

\begin{abstract}

This paper presents a real-time, object-independent grasp synthesis method which can be used for closed-loop grasping.
Our proposed Generative Grasping Convolutional Neural Network (GG-CNN) predicts the quality and pose of grasps at every pixel. This one-to-one mapping from a depth image overcomes limitations of current deep-learning grasping techniques by avoiding discrete sampling of grasp candidates and long computation times.
Additionally, our GG-CNN is orders of magnitude smaller while detecting stable grasps with equivalent performance to current state-of-the-art techniques. 
The light-weight and single-pass generative nature of our GG-CNN allows for closed-loop control at up to 50Hz, enabling accurate grasping in non-static environments where objects move and in the presence of robot control inaccuracies.
In our real-world tests, we achieve an 83\% grasp success rate on a set of previously unseen objects with adversarial geometry and 88\% on a set of household objects that are moved during the grasp attempt.
We also achieve 81\% accuracy when grasping in dynamic clutter.

\end{abstract}

\IEEEpeerreviewmaketitle

\section{Introduction}

In order to perform grasping and manipulation tasks in the unstructured environments of the real world, a robot must be able to compute grasps for the almost unlimited number of objects it might encounter. 
In addition, it needs to be able to act in dynamic environments, whether that be changes in the robot's workspace, noise and errors in perception, inaccuracies in the robot's control, or perturbations to the robot itself.

Robotic grasping has been investigated for decades, yielding a multitude of different techniques~\cite{bicchi2000robotic, bohg2014data, sahbani2012overview, shimoga1996robot}.  Most recently, deep learning techniques have enabled some of the biggest advancements in grasp synthesis for unknown items.  These approaches allow learning of features that correspond to good quality grasps that exceed the capabilities of human-designed features~\cite{Johns2016DeepUncertainty, Lenz2015DeepGrasps, Mahler2017Dex2,Pinto2016SupersizingHours}.  

\begin{figure}[tpb]
    \centering
    \includegraphics[width=\columnwidth]{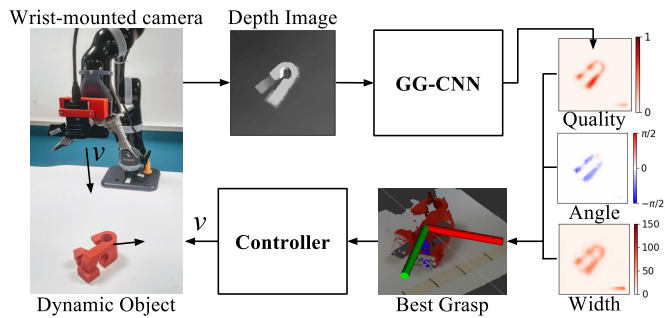}
    \vspace{-5mm}
    \caption{Our real-time, generative grasping pipeline.  A camera mounted to the wrist of the robot captures depth images containing an object to be grasped. Our Generative Grasping Convolutional Neural Network (GG-CNN) generates antipodal grasps -- parameterised as a grasp quality, angle and gripper width -- for every pixel in the input image in a fraction of a second.  The best grasp is calculated and a velocity command ($v$) is issued to the robot. 
    The closed-loop system is capable of grasping dynamic objects and reacting to control errors.
    }
  \label{fig:hero} 
  \vspace{-6mm}
\end{figure}

However, these approaches typically use adapted versions of Convolutional Neural Network (CNN) architectures designed for object recognition~\cite{Johns2016DeepUncertainty,Kumra2017RoboticNetworks, Pinto2016SupersizingHours,Redmon2015Real-timeNetworks}, and in most cases sample and rank grasp candidates individually~\cite{Lenz2015DeepGrasps, Mahler2017Dex2, Pinto2016SupersizingHours}, resulting in long computation times in the order of a second~\cite{Mahler2017Dex2} to tens of seconds~\cite{Lenz2015DeepGrasps}.  As such, these techniques are rarely used in closed-loop grasp execution and rely on precise camera calibration and precise robot control to grasp successfully, even in static environments. 

We propose a different approach to selecting grasp points for previously unseen items.  Our Generative Grasping Convolutional Neural Network (GG-CNN) directly generates an antipodal grasp pose and quality measure for every pixel in an input depth image and is fast enough for closed-loop control of grasping in dynamic environments (\fig{fig:hero}).  We use the term ``generative" to differentiate our direct grasp generation method from methods which sample grasp candidates.

The advantages of GG-CNN over other state-of-the-art grasp synthesis CNNs are twofold.  Firstly, we do not rely on sampling of grasp candidates, but rather directly generate grasp poses on a pixelwise basis, analogous to advances in object detection where fully-convolutional networks are commonly used to perform pixelwise semantic segmentation rather than relying on sliding windows or bounding boxes~\cite{long2015fully}.  
Secondly, our GG-CNN has orders of magnitude fewer parameters than other CNNs used for grasp synthesis, allowing our grasp detection pipeline to execute in only 19\ms~on a GPU-equipped desktop computer, fast enough for closed-loop grasping.

We evaluate the performance of our system in different scenarios by performing grasping trials with a Kinova Mico robot, with static, dynamic and cluttered objects.  In dynamic grasping trials, where objects are moved during the grasp attempt, we achieve 83\% grasping success rate on a set of eight 3D-printed objects with adversarial geometry~\cite{Mahler2017Dex2} and 88\% on a set of 12 household items chosen from standardised object sets.  Additionally, we reproduce the dynamic clutter grasping experiments of~\cite{Viereck2017LearningImages} and show an improved grasp success rate of 81\%.  We further illustrate the advantages of using a closed-loop method by reporting experimental results when artificial inaccuracies are added to the robot's control.

\begin{table*}[t!]
    \begin{center}
        \vspace{-3mm}
        \begin{tabular}{@{}lcccccccccc@{}}
        \toprule
        & \cite{Lenz2015DeepGrasps} & \cite{Redmon2015Real-timeNetworks} & \cite{Pinto2016SupersizingHours} & \cite{Johns2016DeepUncertainty} & \cite{Kumra2017RoboticNetworks} & \cite{Mahler2017Dex2} & \cite{Levine2017LearningCollection} & \cite{Viereck2017LearningImages} & \textbf{Ours} \\
        \midrule
        
        Real Robot Experiments             & \yes & \no  & \yes & \yes & \no  & \yes & \yes & \yes & \yes \\
        Objects from Standard Sets         & \no  & -    & \no  & \no  & -    & \no  & \no  & \no  & \yes \\
        Adversarial Objects~\cite{Mahler2017Dex2}
                                           & \no  & -    & \no  & \no  & -    & \yes & \no  & \no  & \yes \\
        Clutter                            & \no  & \no  & \yes & \no  & \no  & \no  & \yes & \yes & \yes \\
        Closed-loop                        & \no  & -    & \no  & \no  & -    & \no  & \yes & \yes & \yes \\
        Dynamic Objects                    & \no  & -    & \no  & \no  & -    & \no  & \no  & \yes & \yes \\
        Code Available                     & \yes & \no  & \yes & \no  & \no  & \yes & \no  & \no  & \yes* \\
        Training Data Available            & \yes & \yes & \yes & \no  & \yes & \yes & \yes & \no  & \yes \\
        Training Data Type                 & Real & Real & Real & Synthetic 
                                                                       & Real & Synthetic
                                                                                     & Real & Synthetic
                                                                                                   & Real \\
                                           & (Cornell~\cite{Lenz2015DeepGrasps}) 
                                                  & (Cornell) 
                                                         &(Trial)&      & (Cornell)
                                                                              &      &(Trial)&      & (Cornell)\\
        
        \bottomrule
        \end{tabular}
    
        \vspace{2mm}
        \caption{ A comparison of our work to related deep learning approaches to grasp synthesis.  \newline * Code is available at \href{https://github.com/dougsm/ggcnn}{https://github.com/dougsm/ggcnn}}
        \label{tab:comparison}
        \vspace{-8mm}
    \end{center}
    
\end{table*}

\section{Related Work}

\textbf{Grasping Unknown Objects}
Grasp synthesis refers to the formulation of a stable robotic grasp for a given object, which is a topic which has been widely researched resulting in a plethora of techniques.  Broadly, these can be classified into analytic methods and empirical methods~\cite{bohg2014data,sahbani2012overview}.  Analytic methods use mathematical and physical models of geometry, kinematics and dynamics to calculate grasps that are stable~\cite{bicchi2000robotic,Prattichizzo2008Grasping}, but tend to not transfer well to the real world due to the difficultly in modelling physical interactions between a manipulator and an object~\cite{bicchi2000robotic, Rubert2017OnSuccess, sahbani2012overview}.

In contrast, empirical methods focus on using models and experience-based approaches.  Some techniques work with known items, associating good grasp points with an offline database of object models or shapes~\cite{detry2009learning, goldfeder2007grasp, miller2003automatic}, or familiar items, based on object classes~\cite{Saxena2008RoboticVision} or object parts~\cite{el2008handling}, but are unable to generalise to new objects.  

For grasping unknown objects, large advancements have been seen recently with a proliferation of vision-based deep-learning techniques~\cite{Lenz2015DeepGrasps, Mahler2017Dex2, Pinto2016SupersizingHours,Redmon2015Real-timeNetworks, Wang2016RobotNetworks}. Many of these techniques share a common pipeline: classifying grasp candidates sampled from an image or point cloud, then ranking them individually using Convolutional Neural Networks (CNN).  Once the best grasp candidate is determined, a robot executes the grasp open-loop (without any feedback) which requires precise calibration between the camera and the robot, precise control of the robot and a completely static environment.

Execution time is the primary reason that grasps are executed open-loop.  In many cases, deep-learning approaches use large neural networks with millions of parameters~\cite{Johns2016DeepUncertainty, Mahler2017Dex2, Pinto2016SupersizingHours} and process grasp candidates using a sliding window at discrete intervals of offset and rotation~\cite{Lenz2015DeepGrasps, Pinto2016SupersizingHours}, which is computationally expensive and results in grasp planning times in the order of a second~\cite{Mahler2017Dex2} to tens of seconds~\cite{Lenz2015DeepGrasps}.  

Some approaches reduce execution time by pre-processing and pruning the grasp candidates~\cite{Lenz2015DeepGrasps, Wang2016RobotNetworks} or predicting the quality of a discrete set of grasp candidates simultaneously~\cite{Johns2016DeepUncertainty, Pinto2016SupersizingHours}, trading off execution time against the number of grasps which are sampled, but ignoring some potential grasps.  

Instead of sampling grasp candidates, both~\cite{Kumra2017RoboticNetworks} and~\cite{Redmon2015Real-timeNetworks} use a deep CNN to regress a single best grasp pose for an input image. However, these regression methods are liable to output the average of the possible grasps for an object, which itself may not be a valid grasp~\cite{Redmon2015Real-timeNetworks}. 

Similar to our method, \citet{varley2015generating} use a neural network to generate pixelwise heatmaps for finger placement in an image, but still rely on a grasp planner to determine the final grasp pose.

We address the issues of execution time and grasp sampling by directly generating grasp poses for every pixel in an image simultaneously, using a comparatively small neural network.

\textbf{Closed-Loop Grasping}
Closed-loop control of a robot to a desired pose using visual feedback is commonly referred to as visual servoing.  The advantages of visual servoing methods are that they are able to adapt to dynamic environments and do not necessarily require fully accurate camera calibration or position control.  A number of works apply visual servoing directly to grasping applications, with a survey given in~\cite{kragic2002survey}.  However, the nature of visual servoing methods mean that they typically rely on hand-crafted image features for object detection~\cite{kober2012playing, Vahrenkamp2008VisualTasks} or object pose estimation~\cite{Horaud1998VisuallyGrasping}, so do not perform any online grasp synthesis but instead converge to a pre-determined goal pose and are not applicable to unknown objects.  

CNN-based controllers for grasping have very recently been proposed to combine deep learning with closed loop grasping~\cite{Levine2017LearningCollection, Viereck2017LearningImages}.  Rather than explicitly performing grasp synthesis, both systems learn controllers which map potential control commands to the expected quality of or distance to a grasp after execution of the control, requiring many potential commands to be sampled at each time step.  In both cases, the control executes at no more than approximately 5\Hz. While both are closed-loop controllers, grasping in dynamic scenes is only presented in~\cite{Viereck2017LearningImages} and we reproduce these experiments. 

The grasp regression methods~\cite{Kumra2017RoboticNetworks, Redmon2015Real-timeNetworks} report real-time performance, but are not validated with robotic experiments.

\textbf{Benchmarking for Robotic Grasping}
Directly comparing results between robotic grasping experiments is difficult due to the wide range of grasp detection techniques used, the lack of standardisation between object sets, and the limitations of different physical hardware, e.g. robot arms, grippers or cameras.  Many people report grasp success rates on sets of ``household" objects, which vary significantly in the number and types of objects used.

The ACRV Picking Benchmark (APB)~\cite{leitner2017acrvpicking} and the YCB Object Set~\cite{calli2017ycb} define item sets and manipulation tasks, but benchmark on tasks such as warehouse order fulfilment (APB) or table setting and block stacking (YCB) rather than raw grasp success rate as is typically reported.  Additionally, many of the items from these two sets are impractically small, large or heavy for many robots and grippers, so have not been widely adopted for robotic grasping experiments.

We propose a set of 20 reproducible items for testing, comprising comprising 8 3D printed adversarial objects from~\cite{Mahler2017Dex2} and 12 items from the APB and YCB object sets, which we believe provide a wide enough range of sizes, shapes and difficulties to effectively compare results while not excluding use by any common robots, grippers or cameras.

In \tab{tab:comparison} we provide a summary of the recent related work on grasping for unknown objects, and how they compare to our own approach.  This is not intended to be a comprehensive review, but rather to highlight the most relevant work.

\section{Grasp Point Definition}

\begin{figure}[tpb]
    \centering
    \includegraphics[width=\columnwidth]{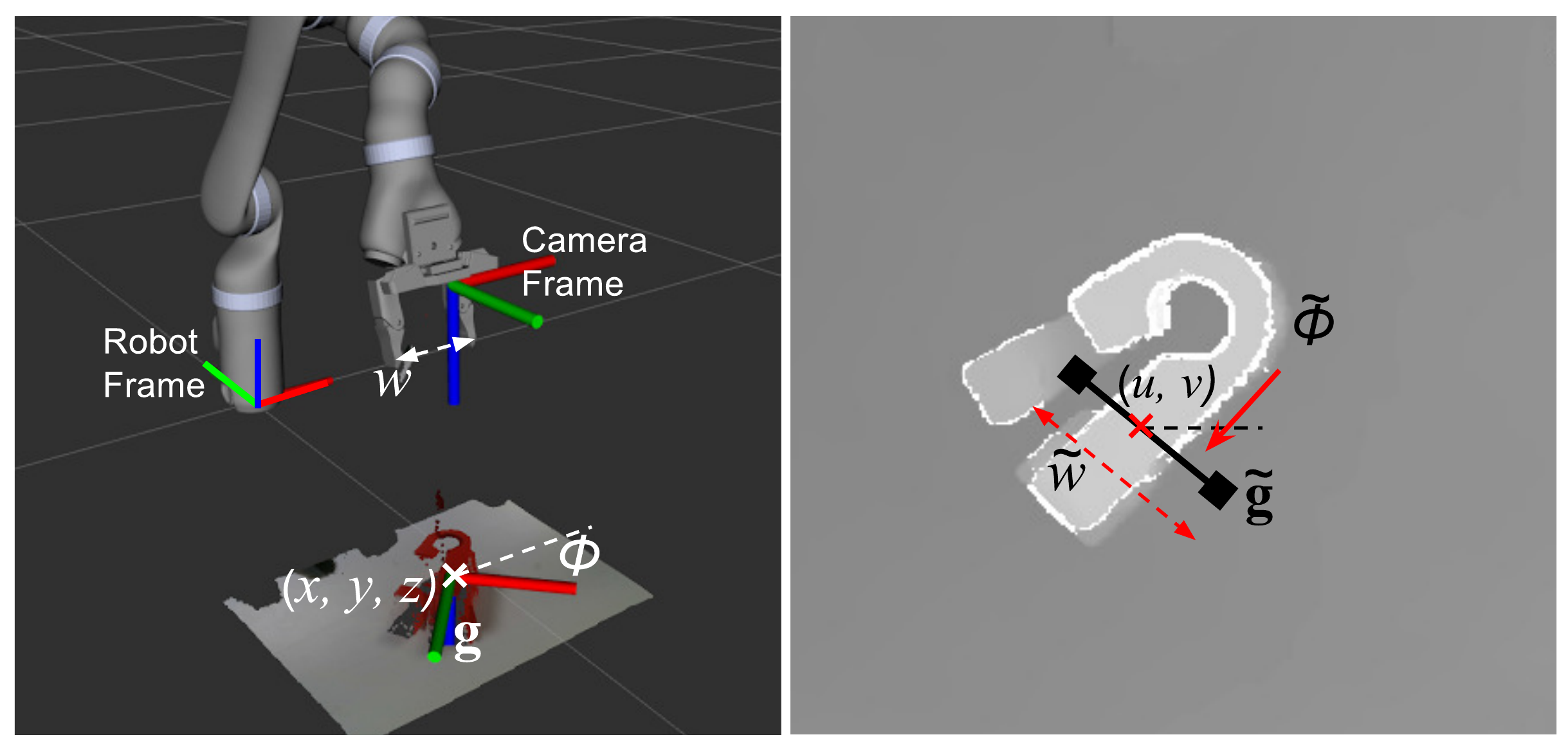}
    \vspace{-6mm}
    \caption{Left: A grasp $\bg$ is defined by its Cartesian position $(x,y,z)$, rotation around the z-axis $\phi$ and gripper width $w$ required for a successful grasp.  Right: In the depth image the grasp pose $\tilde{\bg}$ is defined by its centre pixel $(u, v)$, its rotation\ \ $\tilde{\phi}$ around the image axis and perceived width $\tilde{w}$. 
    }
  \label{fig:problemdefinition} 
  \vspace{-6mm}
\end{figure}

Like much of the related literature \cite{Johns2016DeepUncertainty, Lenz2015DeepGrasps, Mahler2017Dex2, Pinto2016SupersizingHours, Viereck2017LearningImages}, we consider the problem of detecting and executing antipodal grasps on unknown objects, perpendicular to a planar surface, given a depth image of the scene (\fig{fig:problemdefinition}).

Let $\bg = (\bp, \phi, w, q)$ define a grasp, executed perpendicular to the $x$-$y$ plane. 
The grasp is determined by its pose, i.e. the gripper's centre position $\bp = (x, y, z)$ in Cartesian coordinates, the gripper's rotation $\phi$ around the $z$ axis and the required gripper width $w$. 
A scalar quality measure $q$, representing the chances of grasp success, is added to the pose. 
The addition of the gripper width enables a better prediction and better performance over the more commonly used position and rotation only representation. 

We want to detect grasps given a 2.5D depth image $\by = \bbR^{H \times W}$ with height $H$ and width $W$, taken from a camera with known intrinsic parameters.  In the image $\by$ a grasp is described by  
\begin{equation*}
\bgt = (\bs, \tilde{\phi}, \tilde{w}, q),
\end{equation*}
where $\bs = (u, v)$ is the centre point in image coordinates (pixels), $\tilde{\phi}$ is the rotation in the camera's reference frame and $\tilde{w}$ is the grasp width in image coordinates.  A grasp in the image space $\bgt$ can be converted to a grasp in world coordinates $\bg$ by applying a sequence of known transforms,
\begin{equation}\label{eq:transform}
    \bg = t_{RC} ( t_{CI} ( \bgt ))
\end{equation}
where $t_{RC}$  transforms from the camera frame to the world/robot frame and $t_{CI}$ transforms from 2D image coordinates to the 3D camera frame, based on the camera intrinsic parameters and known calibration between the robot and camera. 

We refer to the set of grasps in the image space as the \textit{grasp map}, which we denote
\begin{equation*}
    \cG = (\bPhi, \cW, \cQ) \in \bbR^{3 \times H \times W}
\end{equation*}
where $\bPhi$, $\cW$ and $\cQ$ are each $\in \bbR^{H \times W}$ and contain values of $\tilde{\phi}$, $\tilde{w}$ and $q$ respectively at each pixel $\bs$.

Instead of sampling the input image to create grasp candidates, we wish to directly calculate a grasp $\bgt$ for each pixel in the depth image $\by$.  To do this, we define a function $\fM$ from a depth image to the \textit{grasp map} in the image coordinates: $\fM(\by) = \cG$.  From $\cG$ we can calculate the best visible grasp in the image space $\tilde{\bg}^* = \underset{\cQ}{\text{max }}\cG$, and calculate the equivalent best grasp in world coordinates $\bg^*$ via \eq{eq:transform}.

\section{Generative Grasping Convolutional\newline Neural Network}

We propose the use of a neural network to approximate the complex function $\fM: \by \to \cG$.
$\fM_\theta$ denotes a neural network with $\theta$ being the weights of the network.

We show that $\fM_\theta(\by) = 
(\cQ_\theta, \bPhi_\theta, \cW_\theta ) \approx \fM(\by)$, can be learned with a training set of inputs $\by_T$ and corresponding outputs $\cG_T$ and applying the L2 loss function $\cL$, such that 
\begin{equation*}
    \theta = \underset{\theta}{\text{argmin }}\cL(\cG_T, \fM_\theta(\by_T)).
\end{equation*}

\subsection{Grasp Representation}
$\cG$ estimates the parameters of a set of grasps, executed at the Cartesian point $\bp$, corresponding to each pixel $\bs$.
We represent the \textit{grasp map} $\cG$ as a set of three images, $\cQ$, $\bPhi$ and $\cW$. 
The representations are as follows:

$\cQ$ is an image which describes the quality of a grasp executed at each point $(u,v)$.  The value is a scalar in the range $[0, 1]$ where a value closer to 1 indicates higher grasp quality, i.e.\ higher chance of grasp success.

$\bPhi$ is an image which describes the angle of a grasp to be executed at each point.  Because the antipodal grasp is symmetrical around $\pm\frac{\pi}{2}$ radians, the angles are given in the range $[-\frac{\pi}{2},\frac{\pi}{2}]$.

\vspace{1mm} 
$\cW$ is an image which describes the gripper width of a grasp to be executed at each point.  To allow for depth invariance, values are in the range of $[0, 150]$ pixels, which can be converted to a physical measurement using the depth camera parameters and measured depth.

\begin{figure}[tpb]
    \centering
    \includegraphics[width=\columnwidth]{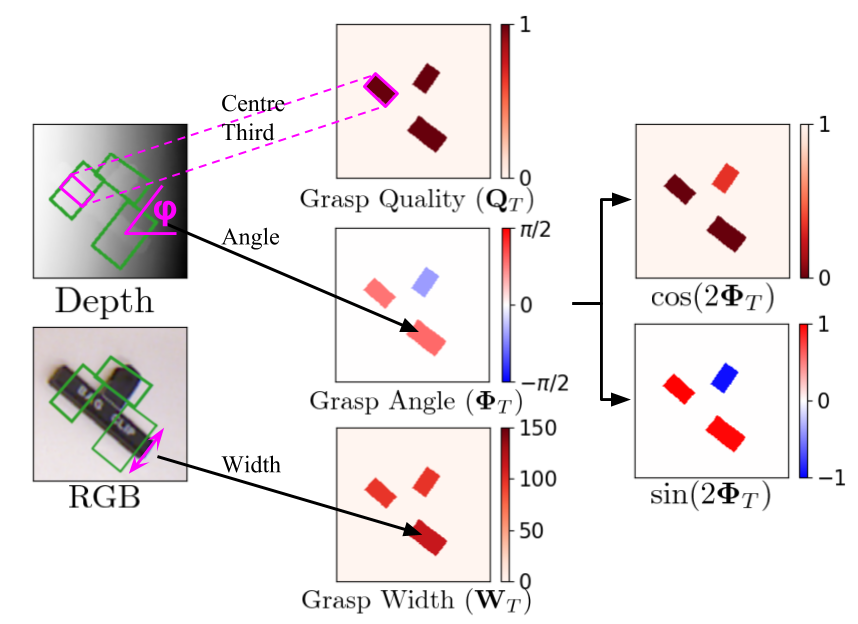}
    \vspace{-6mm}
    \caption{Generation of training data used to train our GG-CNN.  Left: The cropped and rotated depth and RGB images from the Cornell Grasping Dataset~\cite{Lenz2015DeepGrasps}, with the ground-truth positive grasp rectangles representing antipodal grasps shown in green.  The RGB image is for illustration and is not used by our system.  Right: From the ground-truth grasps, we generate the Grasp Quality ($\cQ_T$), Grasp  Angle ($\bPhi_T$) and Grasp Width ($\cW_T$) images to train our network.  The angle is further decomposed into $\cos(2\bPhi_T)$ and $\sin(2\bPhi_T)$ for training as described in \secn{sec:trainingdata}.
    }
  \label{fig:training} 
  \vspace{-6mm}
\end{figure}

\subsection{Training Dataset}
\label{sec:trainingdata}

To train our network, we create a dataset (\fig{fig:training}) from the Cornell Grasping Dataset~\cite{Lenz2015DeepGrasps}.  The Cornell Grasping Dataset contains 885 RGB-D images of real objects, with 5110 human-labelled positive and 2909 negative grasps.  While this is a relatively small grasping dataset compared to some more recent, synthetic datasets~\cite{Mahler2016Dex1, Mahler2017Dex2}, the data best suits our pixelwise grasp representation as multiple labelled grasps are provided per image. 
This is a more realistic estimate of the full pixel-wise \textit{grasp map}, than using a single image to represent one grasp, such as in ~\cite{Mahler2017Dex2}.
We augment the Cornell Grasping Dataset with random crops, zooms and rotations to create a set of 8840 depth images and associated \textit{grasp map} images $\cG_T$, effectively incorporating 51,100 grasp examples.  

The Cornell Grasping Dataset represents antipodal grasps as rectangles using pixel coordinates, aligned to the position and rotation of a gripper~\cite{Jiang2011EfficientRepresentation}.  To convert from the rectangle representation to our image-based representation $\cG$, we use the centre third of each grasping rectangle as an image mask which corresponds to the position of the centre of the gripper.  We use this image mask to update sections of our training images, as described below and shown in \fig{fig:training}.  We consider only the positive labelled grasps for training our network and assume any other area is not a valid grasp.

\textbf{Grasp Quality:}  We treat each ground-truth positive grasp from the Cornell Grasping Dataset as a binary label and set the corresponding area of $\cQ_T$ to a value of 1. All other pixels are 0.

\textbf{Angle:} We compute the angle of each grasping rectangle in the range $[-\frac{\pi}{2},\frac{\pi}{2}]$, and set the corresponding area of $\bPhi_T$.  We encode the angle as two vector components on a unit circle, producing values in the range $[-1, 1]$ and removing any discontinuities that would occur in the data where the angle wraps around $\pm\frac{\pi}{2}$ if the raw angle was used, making the distribution easier for the network to learn~\cite{hara2017designing}.  
Because the antipodal grasp is symmetrical around $\pm\frac{\pi}{2}$ radians, we use use two components $\sin(2\bPhi_T)$ and $\cos(2\bPhi_T)$ which provides values which are unique within $\bPhi_T \in [-\frac{\pi}{2},\frac{\pi}{2}]$ and symmetrical at $\pm\frac{\pi}{2}$.

\textbf{Width:} Similarly, we compute the width in pixels (maximum of 150) of each grasping rectangle representing the width of the gripper and set the corresponding portion of $\cW_T$.  During training, we scale the values of $\cW_T$ by $\frac{1}{150}$ to put it in the range $[0, 1]$.  The physical gripper width can be calculated using the parameters of the camera and the measured depth.

\textbf{Depth Input:} As the Cornell Grasping Dataset is captured with a real camera it already contains realistic sensor noise and therefore no noise addition is required. The depth images are inpainted using OpenCV~\cite{opencv_library} to remove invalid values.  We subtract the mean of each depth image, centring its value around $0$ to provide depth invariance.

\subsection{Network Architecture}

\begin{figure*}[t]
    \centering
    \includegraphics[width=\textwidth]{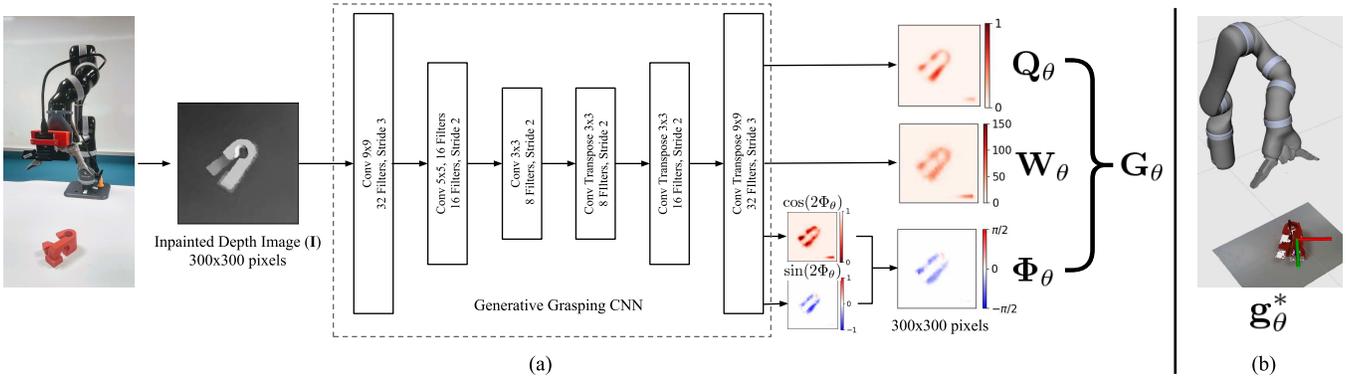}
    \vspace{-8mm}
    \caption{(a) The Generative Grasping CNN (GG-CNN) takes an inpainted depth image ($\by$), and directly generates a grasp pose for every pixel (the \textit{grasp map} $\cG_\theta$), comprising the grasp quality $\cQ_\theta$, grasp width $\cW_\theta$ and grasp angle $\bPhi_\theta$.  (b) From the combined network output, we can compute the best grasp point to reach for, $\bg^*_\theta$.
    }
  \label{fig:architecture} 
  \vspace{-4mm}
\end{figure*}

Our GG-CNN is a fully convolutional topology, shown in \fig{fig:architecture}a. It is used to directly approximate the \textit{grasp map} $\cG_\theta$ from an input depth image $\by$.  Fully convolutional networks have been shown to perform well at computer vision tasks requiring transfer between image domains, such image segmentation~\cite{badrinarayanan2015segnet, long2015fully} and contour detection~\cite{yang2016object}.

The GG-CNN computes the function $\fM_\theta(\by) = (\cQ_\theta, \bPhi_\theta, \cW_\theta )$, where $\by$, $\cQ_\theta$, $\bPhi_\theta$ and $\cW_\theta$ are represented as 300$\times$300 pixel images.  As described in \secn{sec:trainingdata}, the network outputs two images representing the unit vector components of $2\bPhi_\theta$, from which we calculate the grasp angles by $\bPhi_\theta = \frac{1}{2}\arctan\frac{\sin(2\bPhi_\theta)}{\cos(2\bPhi_\theta)}$.

Our final GG-CNN contains 62,420 parameters, making it significantly smaller and faster to compute than the CNNs used for grasp candidate classification in other works which contain hundreds of thousands~\cite{he2016deep, Levine2017LearningCollection} or millions~\cite{Johns2016DeepUncertainty, Mahler2017Dex2, Pinto2016SupersizingHours, Redmon2015Real-timeNetworks} of parameters.  Our code is available at \href{https://github.com/dougsm/ggcnn}{https://github.com/dougsm/ggcnn}.

\subsection{Training}

We train our network on 80\% of our training dataset, and keep 20\% as an evaluation dataset.  We trained 95 networks with similar architectures but different combinations of convolutional filters and stride sizes for 100 epochs each.

To determine the best network configuration, we compare relative performance between our trained networks by evaluating each on detecting ground-truth grasps in our 20\% evaluation dataset containing 1710 augmented images.  

\section{Experimental Set-up}

\subsection{Physical Components}

To perform our grasping trials we use a Kinova Mico 6DOF robot fitted with a Kinova KG-2 2-fingered gripper.

Our camera is an Intel RealSense SR300 RGB-D camera.  The camera is mounted to the wrist of the robot, approximately 80\mm~above the closed fingertips and inclined at $14^\circ$ towards the gripper.  This set-up is shown in \fig{fig:architecture}a.

The GG-CNN computations were performed on a PC running running Ubuntu 16.04 with a 3.6\GHz~Intel Core i7-7700 CPU and NVIDIA GeForce GTX 1070 graphics card.  On this platform, the GG-CNN takes 6\ms~to compute for a single depth image, and computation of the entire grasping pipeline (\secn{sec:pipeline}) takes 19\ms, with the code predominantly written in Python.

\subsubsection{Limitations}
\label{sec:limitations}

The RealSense camera has a specified minimum range of 200\mm.  In reality, we find that the RealSense camera is unable to produce accurate depth measurements from a distance closer than 150\mm, as the separation between the camera's infra-red projector and camera causes shadowing in the depth image caused by the object.  For this reason, when performing closed-loop grasping trials (\secn{sec:closedloop}), we stop updating the target grasp pose at this point, which equates to the gripper being approximately 70\mm~from the object.  Additionally, we find that the RealSense is unable to provide any valid depth data on many black or reflective objects.

The Kinova KG-2 gripper has a maximum stroke of 175\mm, which could easily envelop many of the test items. To encourage more precise grasps, we limit the maximum gripper width to approximately 70\mm.  The fingers of the gripper have some built-in compliance and naturally splay slightly at the tips, so we find that objects with a height less than 15\mm~(especially those that are cylindrical, like a thin pen) cannot be grasped.

\subsection{Test Objects}
\label{sec:objects}

\begin{figure}[tpb]
    \centering
    \includegraphics[width=\columnwidth]{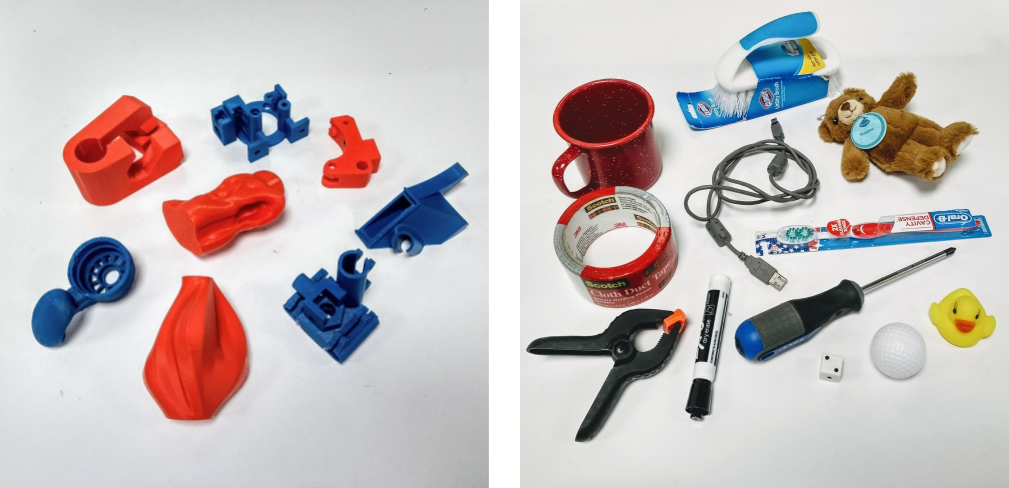}
    \vspace{-6mm}
    \caption{The objects used for grasping experiments. Left: The 8 adversarial objects from~\cite{Mahler2017Dex2}. Right: The 12 household objects selected from~\cite{calli2017ycb} and~\cite{leitner2017acrvpicking}.
    }
  \label{fig:objects} 
  \vspace{-6mm}
\end{figure}

There is no set of test objects which are commonly used for robotic grasping experiments, with many people using random ``household" objects which are not easily reproducible.  We propose here two sets of reproducible benchmark objects (\fig{fig:objects}) on which we test the grasp success rate of our approach.

\textbf{Adversarial Set} The first set consists of eight 3D-printed objects with adversarial geometry, which were used by \citet{Mahler2017Dex2} to verify the performance of their Grasp Quality CNN.  The objects all have complex geometry, meaning there is a high chance of a collision with the object in the case of an inaccurate grasp, as well as many curved and inclined surfaces which are difficult or impossible to grasp.  The object models are available online as part of the released datatasets for Dex-Net 2.0\footnote{\href{https://berkeleyautomation.github.io/dex-net/}{https://berkeleyautomation.github.io/dex-net/}}~\cite{Mahler2017Dex2}.

\textbf{Household Set} This set of items contains twelve household items of varying sizes, shapes and difficulty with minimal redundancy (i.e. minimal objects with similar shapes).  The objects were chosen from the standard robotic grasping datasets the ACRV Picking Benchmark (APB)~\cite{leitner2017acrvpicking} and the YCB Object Set~\cite{calli2017ycb}, both of which provide item specifications and online purchase links.  Half of the item classes (mug, screwdriver, marker pen, die, ball and clamp) appear in both data sets.  We have made every effort to produce a balanced object set containing objects which are deformable (bear and cable), perceptually challenging (black clamp and screwdriver handle, thin reflective edges on the mug and duct tape, and clear packaging on the toothbrush), and objects which are small and require precision (golf ball, duck and die). 

While both the APB and YCB object sets contain a large number of objects, many are physically impossible for our robot to grasp due to being too small and thin (e.g. screws, washers, envelope), too large (e.g. large boxes, saucepan, soccer ball) or too heavy (e.g. power drill, saucepan).
While manipulating these objects is an open problem in robotics, we do not consider them for our experiments in order to compare our results to other work which use similar object classes to ours~\cite{Johns2016DeepUncertainty, Lenz2015DeepGrasps,Levine2017LearningCollection, Mahler2017Dex2, Pinto2016SupersizingHours}.

\subsection{Grasp Detection Pipeline}
\label{sec:pipeline}

Our grasp detection pipeline comprises three stages: image processing, evaluation of the GG-CNN and computation of a grasp pose.

The depth image is first cropped to a square, and scaled to $300\times300$ pixels to suit the input of the GG-CNN.  We inpaint invalid depth values using OpenCV~\cite{opencv_library}. 

The GG-CNN is then evaluated on the processed depth image, to produce the \textit{grasp map} $\cG_\theta$.  We filter $\cQ_\theta$ with a Gaussian kernel, similar to ~\cite{Johns2016DeepUncertainty}, and find this helps improve our grasping performance by removing outliers and causing the local maxima of $\cG_\theta$ to converge to regions of more robust grasps.

Finally, the best grasp pose in the image space $\bgt^*_\theta$ is computed by identifying the maximum pixel $\bs^*$ in $\cQ_\theta$, and the rotation and width are computed from $\bPhi_\theta |_{\bs^*}$ and $\cW_\theta |_{\bs^*}$ respectively.  The grasp in Cartesian coordinates $\bg^*_\theta$ is computed via \eq{eq:transform} (\fig{fig:architecture}b).

\subsection{Grasp Execution}

We evaluate the performance of our system using two grasping methods.  Firstly, an open-loop grasping method similar to~\cite{Lenz2015DeepGrasps, Pinto2016SupersizingHours, Mahler2017Dex2}, where the best grasp pose is calculated from a single viewpoint and executed by the robot open-loop.  Secondly, we implement a closed-loop visual servoing controller which we use for evaluating our system in dynamic environments.

\subsubsection{Open Loop Grasping}
\label{sec:openloop}

To perform open-loop grasps, the camera is positioned approximately 350\mm~above and parallel to the surface of the table.  An item is placed in the field of view of the camera.  A depth image is captured and the pose of the best grasp is computed using the grasp detection pipeline.  The robot moves to a pre-grasp position, with the gripper tips aligned with and approximately 170\mm~above the computed grasp.  From here, the robot moves straight down until the grasp pose is met or a collision is detected via force feedback in the robot.  The gripper is closed and lifted, and the grasp is recorded as a success if the object is successfully lifted to the starting position.

\subsubsection{Closed Loop Grasping}
\label{sec:closedloop}

To perform closed-loop grasping, we implement a Position Based Visual Servoing (PBVS) controller~\cite{kragic2002survey}.  The camera is initially positioned approximately 400\mm~above the surface of the table, and an object is placed in the field of view.  Depth images are generated at a rate of 30\Hz~and processed by the grasp detection pipeline to generate grasp poses in real time.  There may be multiple similarly-ranked good quality grasps in an image, so to avoid rapidly switching between them, which would confuse the controller, we compute three grasps from the highest local maxima of $\cG_\theta$ and select the one which is closest (in image coordinates) to the grasp  used on the previous iteration.  As the control loop is fast compared to the movement of the robot, there is unlikely to be a major change between frames.  The system is initialised to track the global maxima of $\cQ_\theta$ at the beginning of each grasp attempt.  We represent the poses of the grasp $\bT_{\bg^*_\theta}$ and the gripper fingers $\bT_{f}$ as 6D vectors comprising the Cartesian position and roll, pitch and yaw Euler angles $(x, y, z, \alpha, \beta, \gamma)$, and generate a 6D velocity signal for the end-effector:

\begin{equation*}
    \mathbf{v} = \mathbf{\lambda} (\bT_{\bg^*_\theta} - \bT_{f})
\end{equation*}
where $\lambda$ is a 6D scale for the velocity, which causes the gripper to converge to the grasp pose.  Simultaneously, we control the gripper fingers to the computed gripper width via velocity control.  Control is stopped when the grasp pose is reached or a collision is detected.  The gripper is closed and lifted and the grasp is recorded as a success if the object is successfully lifted to the starting position.

\subsection{Object Placement}

To remove bias related to object pose, objects are shaken in a cardboard box and emptied into the robot's workspace for each grasp attempt.  The workspace is an approximately 250$\times$300\mm~area in the robot's field of view in which the robot's kinematics allow it to execute a vertical grasp.

\section{Experiments}

\begin{table*}[t!]
    \begin{center}
        \vspace{-3mm}
        \begin{tabular}{@{}lcccccccc@{}}
        \toprule
        & \cite{Lenz2015DeepGrasps} & \cite{Pinto2016SupersizingHours} & \cite{Johns2016DeepUncertainty} & \cite{Mahler2017Dex2} & \cite{Levine2017LearningCollection} & \cite{Viereck2017LearningImages} & \textbf{Ours} \\
        \midrule
        
        \textbf{Grasp Success Rate (\%)} \\                                    
        Household Objects (Static)$^\#$   & 89    & 73    & 80   & 80   & 80  &      & 92$\pm$5  \\
        Adversarial Objects (Static)      &       &       &      & \textbf{93}*& &   & 84$\pm$8 \\
        Household Objects (Dynamic)       &       &       &      &      &     &      & \textbf{88}$\pm$6 \\
        Adversarial Objects (Dynamic)     &       &       &      &      &     &      & \textbf{83}$\pm$8 \\
        Objects from~\cite{Viereck2017LearningImages} (Single)
                                          &       &       &      &      &     & 98   & \textbf{100} \\
        Objects from~\cite{Viereck2017LearningImages} (Clutter)
                                          &       &       &      &      &     & \textbf{89} & 87$\pm$7 \\
        Objects from~\cite{Viereck2017LearningImages} (Clutter, Dynamic)
                                          &       &       &      &      &     & 77   & \textbf{81}$\pm$8 \\
                                             
        \midrule
        Network Parameters (approx.)      &       & 60M   & 60M  & 18M  & 1M  &      & \textbf{62k} \\
        Computation Time (to generate pose or command) & 13.5s &       &      & 0.8s & 0.2-0.5s & 0.2s & \textbf{19ms} \\
        \bottomrule
        \end{tabular}
    
        \vspace{2mm}
        \caption{Results from grasping experiments with 95\% confidence intervals, and comparison to other deep learning approaches where available. \newline \# Note that all experiments use different item sets and experimental protocol, so comparative performance is indicative only. \newline *Contrary to our approach, \cite{Mahler2017Dex2} train their grasp network on the adversarial objects! }
        \label{tab:results}
        \vspace{-8mm}
    \end{center}
    
\end{table*}

To evaluate the performance of our grasping pipeline and GG-CNN, we perform several experiments comprising over 2000 grasp attempts.  In order to compare our results to others, we aim to reproduce similar experiments where possible, and also aim to present experiments which are reproducible in themselves by using our defined set of objects (\secn{sec:objects}) and defined dynamic motions.  

Firstly, to most closely compare to existing work in robotic grasping, we perform grasping on singulated, static objects from our two object sets.  Secondly, to highlight our primary contribution, we evaluate grasping on objects which are moved during the grasp attempt, to show the ability of our system to perform dynamic grasping.  Thirdly, we show our system's ability to generalise to dynamic cluttered scenes by reproducing the experiments from~\cite{Viereck2017LearningImages} and show improved results.  Finally, we further show the advantage of our closed-loop grasping method over open-loop grasping by performing grasps in the presence of simulated kinematic errors of our robot's control.  

\tab{tab:results} provides a summary of our results in different grasping tasks and comparisons to other work where possible.

\subsection{Static Grasping}
\label{sec:baseline}

To evaluate the performance of our GG-CNN under static conditions, we performed grasping trials using both the open- and closed-loop methods on both sets of test objects, using the set-up shown in \fig{fig:dynamic}a. We perform 10 trials on each object.  For the adversarial object set, the grasp success rates were 84\% (67/80) and 81\% (65/80) for the open- and closed-loop methods respectively.  For the household object set, the open-loop method achieved 92\% (110/120) and the closed-loop 91\% (109/120).  

A comparison to other work is provided in \tab{tab:results}.  We note that the results may not be directly comparable due to the different objects and experimental protocol used, however we aim to show that we achieve comparable performance to other works which use much larger neural networks and have longer computation times. A noteworthy difference in method is \cite{Levine2017LearningCollection}, which does not require precise camera calibration, but rather learns the spatial relationship between the robot and the objects using vision.

\subsection{Dynamic Grasping}
\label{sec:dynamic}

\begin{figure}[tpb]
    \centering
    \includegraphics[width=\columnwidth]{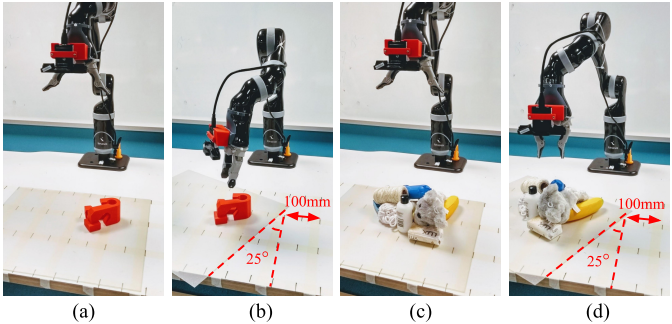}
    \vspace{-8mm}
    \caption{Grasping experiments. (a) Set-up for static grasping, and initial set-up for dynamic grasping. (b) During a dynamic grasp attempt, the object is translated at least 100\mm~and rotated at least 25$^\circ$, measured by the grid on the table.  (c) Set-up for static grasping in clutter, and initial set-up for dynamic grasping in clutter.  (d) During a dynamic grasp attempt, the cluttered objects are translated at least 100\mm~and rotated at least 25$^\circ$, measured by the grid on the table.
    }
    \label{fig:dynamic} 
    \vspace{-6mm}
\end{figure}

To perform grasps on dynamic objects we take inspiration from recent work in~\cite{Viereck2017LearningImages}, where items are moved once by hand randomly during each grasp attempt.  To assist reproducibility, we define this movement to consist of a translation of at least 100\mm~ and a rotation of at least 25$^\circ$ after the grasp attempt has begun, shown in \fig{fig:dynamic}a-b, which we measure using a grid on the table.    

We perform 10 grasp attempts on each adversarial and household object using our closed-loop method, and achieve grasp success rates of 83\% (66/80) for the adversarial objects and 88\% (106/120) for the household objects.  These results are not significantly different to our results on static objects, and are within the 95\% confidence bounds of our results on static objects, showing our method's ability to maintain a high level of accuracy when grasping dynamic objects.  

We do not compare directly to an open-loop method as the object movement moves the object sufficiently far from the original position that no successful grasps would be possible.

\subsection{Dynamic Grasping in Clutter}
\label{sec:clutter}

\begin{figure}[tpb]
    \centering
    \includegraphics[width=\columnwidth]{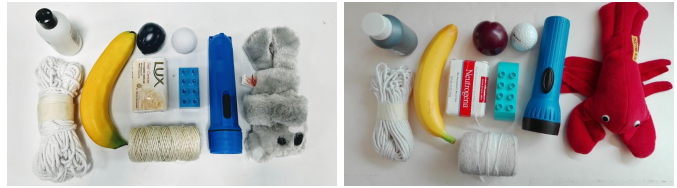}
    \vspace{-6mm}
    \caption{Left: The objects used to reproduce the dynamic grasping in clutter experiment of~\cite{Viereck2017LearningImages}.  Right: The test objects used by~\cite{Viereck2017LearningImages}. We have attempted to recreate the object set as closely as possible.}
    \label{fig:comparison} 
    \vspace{-6mm}
\end{figure}

\citet{Viereck2017LearningImages} demonstrate a visuomotor controller for robotic grasping in clutter that is able to react to disturbances to the objects being grasped.  As this work is closely related to our own, we have made an effort to recreate their experiments using objects as close as possible to their set of 10 (\fig{fig:comparison}) to perform a comparison.  Even though our GG-CNN has not been trained on cluttered environments, we show here its ability to perform grasping in the presence of clutter.  We recreate the three grasping experiments from~\cite{Viereck2017LearningImages} as follows:

\subsubsection{Isolated Objects}

We performed 4 grasps on each of the 10 test objects (\fig{fig:comparison}) in isolation, and achieved a grasp success rate of 100\%, compared to 98\% (39/40) in \cite{Viereck2017LearningImages}.

\subsubsection{Cluttered Objects}

The 10 test objects are shaken in a box and emptied in a pile below the robot (\fig{fig:dynamic}c).  The robot attempts multiple grasps, and any objects that are grasped are removed.  This continues until all objects are grasped, three consecutive grasps are failures or all objects are outside the workspace of the robot.  We run this experiment 10 times.

Despite our GG-CNN not being trained on cluttered scenes, we achieved a grasp success rate of 87\% (83/96) compared to 89\% (66/74) in~\cite{Viereck2017LearningImages}.  Our most common failure cause was collision of the gripper with two objects that had fallen up against each other.  8 out of the 13 failed grasps were from two runs where objects had fallen into an ungraspable position and failed repeatedly.  8 out of the 10 runs finished with 0 or 1 grasp failures.

\subsubsection{Dynamic Cluttered Objects}

For dynamic scenes, we repeat the procedure as above with the addition of a random movement of the objects during the grasp attempt.  \citet{Viereck2017LearningImages} do not give specifications for their random movement, so we use the same procedures as in \secn{sec:dynamic}, where we move the objects randomly, at least 100\mm~and 25$^\circ$ during each grasp attempt (\fig{fig:dynamic}d).

In 10 runs of the experiment, we performed 94 grasp attempts of which 76 were successful (81\%), compared to 77\% (58/75) in~\cite{Viereck2017LearningImages}.  Like the static case, 8 of the 18 failed grasps were from two runs where the arrangement of the objects resulted in repeated failed attempts.  In the other 8 runs, all available objects (i.e. those that didn't fall/roll out of the workspace) were successfully grasped with 2 or fewer failed grasps.  

Despite not being trained on cluttered scenes, this shows our approach's ability to perform grasping in clutter and its ability to react to dynamic scenes, showing only a 5\% decrease in performance for the dynamic case compared to 12\% in~\cite{Viereck2017LearningImages}.

For the same experiments, ~\cite{Viereck2017LearningImages} shows that an open-loop baseline approach on the same objects that is able to achieve 95\% grasp success rate for the static cluttered scenes achieves only 23\% grasp success rate for dynamic scenes as it is able to react to the change in item location.

\subsection{Robustness to Control Errors}
\label{sec:controlerrors}

The control of a robot may not always be precise. For example, when performing grasping trials with a Baxter Research Robot, \citet{Lenz2015DeepGrasps} found that positioning errors of up to 20\mm~were typical.  A major advantage of using a closed-loop controller for grasping is the ability to perform accurate grasps despite inaccurate control.  We show this by simulating an inaccurate kinematic model of our robot by introducing a cross-correlation between Cartesian ($x$, $y$ and $z$) velocities:

\begin{equation*}
    \mathbf{v}_c = \mathbf{v} \cdot \begin{pmatrix} 1 + c_{xx} & c_{xy} & c_{xz} \\ c_{yx} & 1 + c_{yy} & c_{yz} \\ c_{zx} & c_{zy} & 1 + c_{zz} \end{pmatrix}
\end{equation*}
where each $c \sim \mathcal{N}(0, \sigma^{2})$ is sampled at the beginning of each grasp attempt.  While a real kinematic error (e.g. a link length being incorrectly configured) would result in a more non-linear response, our noise model provides a good approximation which is independent of the robot's kinematic model, so has a deterministic effect with respect to end-effector positioning and is more easily replicated on a different robotic system.  

We test grasping on both object sets with 10 grasp attempts per object for both the open- and closed-loop methods with $\sigma =$ 0.0 (the baseline case), 0.05, 0.1 and 0.15.  In the case of our open-loop controller, where we only control velocity for 170\mm~in the $z$ direction from the pre-grasp pose (\secn{sec:openloop}), this corresponds to having a robot with an end-effector precision described by a normal distribution with zero mean and standard deviation 0.0, 8.5, 17.0 and 25.5\mm~respectively, by the relationship for scalar multiplication of the normal distribution:
\begin{equation*}
    \Delta x = \Delta y = \Delta z \cdot \mathcal{N}(0, \sigma^2) = \mathcal{N}(0, \Delta z^2 \sigma^2) ; \Delta z = \text{170\mm}
\end{equation*}

\begin{figure}[tpb]
    \centering
    \includegraphics[width=\columnwidth]{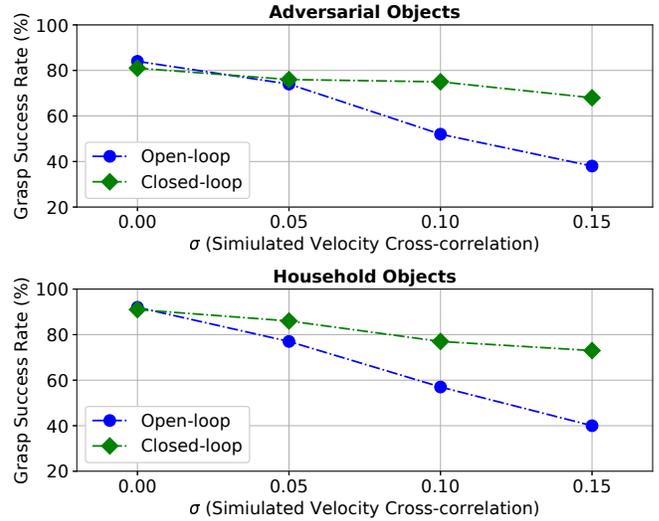}
    \vspace{-8mm}
    \caption{Comparison of grasp success rates for open-loop and closed-loop control methods with velocity cross-correlation added to simulate kinematic errors (see \secn{sec:controlerrors} for full details).  The closed-loop method out-performs the open-loop method in all cases where kinematic errors are present.  10 trials were performed on each object in both the adversarial and household object sets.
    }
  \label{fig:kinematicerror} 
  \vspace{-6mm}
\end{figure}

The results are illustrated in \fig{fig:kinematicerror}, and show that the closed-loop method outperforms the open-loop method in the presence of control error.  This highlights a major advantage of being able to perform closed-loop grasping, as the open-loop methods are unable to respond, achieving only 38\% grasp success rate in the worst case.  In comparison, the closed-loop method achieves 68\% and 73\% grasp success rate in the worst case on the adversarial and household objects respectively.

The decrease in performance of the closed-loop method is due to the limitation of our camera (\secn{sec:limitations}), where we are unable to update the grasp pose when the gripper is within 70\mm~of the object, so can not correct for errors in this range.

The addition of control inaccuracy effects objects which require precise grasps (e.g. the adversarial objects, and small objects such as the die and ball) the most.  Simpler objects which are more easily caged by the gripper, such as the pen, still report good grasp results in the presence of kinematic error.

\section{Conclusion} 
\label{sec:conclusion}

We present our Generative Grasping Convolutional Neural Network (GG-CNN), an object-independent grasp synthesis model which directly generates grasp poses from a depth image on a pixelwise basis, instead of sampling and classifying individual grasp candidates like other deep learning techniques.  Our GG-CNN is orders of magnitude smaller than other recent grasping networks, allowing us to generate grasp poses at a rate of up to 50\Hz~and perform closed-loop control.  
We show through grasping trials that our system is able to gain state-of-the-art results in grasping unknown, dynamic objects, including objects in dynamic clutter.  
Additionally, our closed-loop grasping method significantly outperforms an open-loop method in the presence of simulated robot control error.

We encourage reproducibility in robotic grasping experiments by using two standard object sets, a set of eight 3D-printed objects with adversarial geometry \cite{Mahler2017Dex2} plus a proposed set of twelve household items from standard robotic benchmark object sets, and by defining the parameters of our dynamic grasping experiments.  On our two object sets we achieve 83\% and 88\% grasp success rate respectively when objects are moved during the grasp attempt, and 81\% for objects in dynamic clutter.

\section*{Acknowledgments}

This research was supported by the Australian Research Council Centre of Excellence for Robotic Vision (project number CE140100016).

\bibliographystyle{plainnat}
\bibliography{references}

\begin{thebibliography}{35}
\providecommand{\natexlab}[1]{#1}
\providecommand{\url}[1]{\texttt{#1}}
\expandafter\ifx\csname urlstyle\endcsname\relax
  \providecommand{\doi}[1]{doi: #1}\else
  \providecommand{\doi}{doi: \begingroup \urlstyle{rm}\Url}\fi

\bibitem[Badrinarayanan et~al.(2015)Badrinarayanan, Kendall, and
  Cipolla]{badrinarayanan2015segnet}
Vijay Badrinarayanan, Alex Kendall, and Roberto Cipolla.
\newblock \href{https://arxiv.org/abs/1511.00561}{SegNet: A Deep Convolutional
  Encoder-Decoder Architecture for Image Segmentation}.
\newblock \emph{arXiv preprint arXiv:1511.00561}, 2015.

\bibitem[Bicchi and Kumar(2000)]{bicchi2000robotic}
Antonio Bicchi and Vijay Kumar.
\newblock \href{http://ieeexplore.ieee.org/abstract/document/844081/}{Robotic
  Grasping and Contact: A Review }.
\newblock In \emph{Proc.\ of the IEEE International Conference on Robotics and
  Automation (ICRA)}, pages 348--353, 2000.

\bibitem[Bohg et~al.(2014)Bohg, Morales, Asfour, and Kragic]{bohg2014data}
Jeannette Bohg, Antonio Morales, Tamim Asfour, and Danica Kragic.
\newblock
  \href{http://ieeexplore.ieee.org/abstract/document/6672028/}{Data-Driven
  Grasp Synthesis -- A Survey}.
\newblock \emph{IEEE Transactions on Robotics}, 30\penalty0 (2):\penalty0
  289--309, 2014.

\bibitem[Bradski(2000)]{opencv_library}
G.~Bradski.
\newblock {The OpenCV Library}.
\newblock \emph{Dr. Dobb's Journal of Software Tools}, 2000.

\bibitem[Calli et~al.(2015)Calli, Walsman, Singh, Srinivasa, Abbeel, and
  Dollar]{calli2017ycb}
Berk Calli, Aaron Walsman, Arjun Singh, Siddhartha Srinivasa, Pieter Abbeel,
  and Aaron~M Dollar.
\newblock \href{http://ieeexplore.ieee.org/document/7254318/}{Benchmarking in
  Manipulation Research: Using the Yale-CMU-Berkeley Object and Model Set}.
\newblock \emph{IEEE Robotics \& Automation Magazine}, 22\penalty0
  (3):\penalty0 36--52, 2015.

\bibitem[Detry et~al.(2009)Detry, Baseski, Popovic, Touati, Kruger, Kroemer,
  Peters, and Piater]{detry2009learning}
Renaud Detry, Emre Baseski, Mila Popovic, Younes Touati, N~Kruger, Oliver
  Kroemer, Jan Peters, and Justus Piater.
\newblock \href{http://ieeexplore.ieee.org/document/5175520/}{Learning
  Object-specific Grasp Affordance Densities}.
\newblock In \emph{Proc.\ of the IEEE International Conference on Development
  and Learning (ICDL)}, pages 1--7, 2009.

\bibitem[El-Khoury and Sahbani(2008)]{el2008handling}
Sahar El-Khoury and Anis Sahbani.
\newblock
  \href{https://infoscience.epfl.ch/record/168926/files/WS_IROS2008.pdf}{Handling
  Objects By Their Handles}.
\newblock In \emph{IEEE/RSJ International Conference on Intelligent Robots and
  Systems (IROS)}, 2008.

\bibitem[Goldfeder et~al.(2007)Goldfeder, Allen, Lackner, and
  Pelossof]{goldfeder2007grasp}
Corey Goldfeder, Peter~K Allen, Claire Lackner, and Raphael Pelossof.
\newblock \href{http://ieeexplore.ieee.org/abstract/document/4209818/}{Grasp
  Planning via Decomposition Trees}.
\newblock In \emph{Proc.\ of the IEEE International Conference on Robotics and
  Automation (ICRA)}, pages 4679--4684, 2007.

\bibitem[Hara et~al.(2017)Hara, Vemulapalli, and Chellappa]{hara2017designing}
Kota Hara, Raviteja Vemulapalli, and Rama Chellappa.
\newblock \href{https://arxiv.org/abs/1702.01499}{Designing Deep Convolutional
  Neural Networks for Continuous Object Orientation Estimation}.
\newblock \emph{arXiv preprint arXiv:1702.01499}, 2017.

\bibitem[He et~al.(2016)He, Zhang, Ren, and Sun]{he2016deep}
Kaiming He, Xiangyu Zhang, Shaoqing Ren, and Jian Sun.
\newblock \href{http://ieeexplore.ieee.org/abstract/document/7780459/}{Deep
  Residual Learning for Image Recognition}.
\newblock In \emph{Proc.\ of the IEEE Conference on Computer Vision and Pattern
  Recognition (CVPR)}, pages 770--778, 2016.

\bibitem[Horaud et~al.(1998)Horaud, Dornaika, and
  Espiau]{Horaud1998VisuallyGrasping}
Radu Horaud, Fadi Dornaika, and Bernard Espiau.
\newblock \href{http://ieeexplore.ieee.org/document/704214/}{Visually Guided
  Object Grasping}.
\newblock \emph{IEEE Transactions on Robotics and Automation}, 14\penalty0
  (4):\penalty0 525--532, 1998.

\bibitem[Johns et~al.(2016)Johns, Leutenegger, and
  Davison]{Johns2016DeepUncertainty}
Edward Johns, Stefan Leutenegger, and Andrew~J. Davison.
\newblock \href{http://ieeexplore.ieee.org/document/7759657/}{{Deep Learning a
  Grasp Function for Grasping under Gripper Pose Uncertainty}}.
\newblock In \emph{Proc.\ of the IEEE/RSJ International Conference on
  Intelligent Robots and Systems (IROS)}, pages 4461--4468, 2016.

\bibitem[Kober et~al.(2012)Kober, Glisson, and Mistry]{kober2012playing}
Jens Kober, Matthew Glisson, and Michael Mistry.
\newblock
  \href{https://pdfs.semanticscholar.org/e4ad/1d8d17aeecd5a6005c3583632d28f63c87af.pdf}{Playing
  Catch and Juggling with a Humanoid Robot}.
\newblock In \emph{Proc.\ of the IEEE-RAS International Conference on Humanoid
  Robots (Humanoids)}, pages 875--881, 2012.

\bibitem[Kragic et~al.(2002)Kragic, Christensen, et~al.]{kragic2002survey}
Danica Kragic, Henrik~I Christensen, et~al.
\newblock
  \href{http://citeseerx.ist.psu.edu/viewdoc/summary?doi=10.1.1.24.1025}{Survey
  on Visual Servoing for Manipulation}.
\newblock \emph{Computational Vision and Active Perception Laboratory,
  Fiskartorpsv}, 2002.

\bibitem[Kumra and Kanan(2017)]{Kumra2017RoboticNetworks}
S.~Kumra and C.~Kanan.
\newblock \href{http://ieeexplore.ieee.org/document/8202237/}{Robotic Grasp
  Detection using Deep Convolutional Neural Networks}.
\newblock In \emph{Proc.\ of the IEEE/RSJ International Conference on
  Intelligent Robots and Systems (IROS)}, pages 769--776, 2017.

\bibitem[Leitner et~al.(2017)Leitner, Tow, S{\"u}nderhauf, Dean, Durham,
  Cooper, Eich, Lehnert, Mangels, McCool, et~al.]{leitner2017acrvpicking}
J{\"u}rgen Leitner, Adam~W Tow, Niko S{\"u}nderhauf, Jake~E Dean, Joseph~W
  Durham, Matthew Cooper, Markus Eich, Christopher Lehnert, Ruben Mangels,
  Christopher McCool, et~al.
\newblock \href{http://ieeexplore.ieee.org/document/7989545/}{The ACRV Picking
  Benchmark: A Robotic Shelf Picking Benchmark to Foster Reproducible
  Research}.
\newblock In \emph{Proc.\ of the IEEE International Conference on Robotics and
  Automation (ICRA)}, pages 4705--4712, 2017.

\bibitem[Lenz et~al.(2015)Lenz, Lee, and Saxena]{Lenz2015DeepGrasps}
Ian Lenz, Honglak Lee, and Ashutosh Saxena.
\newblock \href{http://journals.sagepub.com/doi/10.1177/0278364914549607}{Deep
  learning for detecting robotic grasps}.
\newblock \emph{The International Journal of Robotics Research (IJRR)},
  34\penalty0 (4-5):\penalty0 705--724, 2015.

\bibitem[Levine et~al.(2016)Levine, Pastor, Krizhevsky, and
  Quillen]{Levine2017LearningCollection}
Sergey Levine, Peter Pastor, Alex Krizhevsky, and Deirdre Quillen.
\newblock \href{http://dx.doi.org/10.1007/978-3-319-50115-4_16}{Learning
  Hand-Eye Coordination for Robotic Grasping with Large-Scale Data Collection}.
\newblock In \emph{International Symposium on Experimental Robotics}, pages
  173--184, 2016.

\bibitem[Long et~al.(2015)Long, Shelhamer, and Darrell]{long2015fully}
Jonathan Long, Evan Shelhamer, and Trevor Darrell.
\newblock \href{http://ieeexplore.ieee.org/document/7478072/}{Fully
  Convolutional Networks for Semantic Segmentation}.
\newblock In \emph{Proc.\ of the IEEE Conference on Computer Vision and Pattern
  Recognition (CVPR)}, pages 3431--3440, 2015.

\bibitem[Mahler et~al.(2016)Mahler, Pokorny, Hou, Roderick, Laskey, Aubry,
  Kohlhoff, Kroger, Kuffner, and Goldberg]{Mahler2016Dex1}
Jeffrey Mahler, Florian~T. Pokorny, Brian Hou, Melrose Roderick, Michael
  Laskey, Mathieu Aubry, Kai Kohlhoff, Torsten Kroger, James Kuffner, and Ken
  Goldberg.
\newblock \href{http://ieeexplore.ieee.org/document/7487342/}{Dex-Net 1.0: A
  cloud-based network of 3D objects for robust grasp planning using a
  Multi-Armed Bandit model with correlated rewards}.
\newblock In \emph{Proc.\ of the IEEE International Conference on Robotics and
  Automation (ICRA)}, pages 1957--1964, 2016.

\bibitem[Mahler et~al.(2017)Mahler, Liang, Niyaz, Laskey, Doan, Liu, Ojea, and
  Goldberg]{Mahler2017Dex2}
Jeffrey Mahler, Jacky Liang, Sherdil Niyaz, Michael Laskey, Richard Doan, Xinyu
  Liu, Juan~Aparicio Ojea, and Ken Goldberg.
\newblock \href{https://arxiv.org/abs/1703.09312}{Dex-Net 2.0: Deep Learning to
  Plan Robust Grasps with Synthetic Point Clouds and Analytic Grasp Metrics}.
\newblock In \emph{Robotics: Science and Systems (RSS)}, 2017.

\bibitem[Miller et~al.(2003)Miller, Knoop, Christensen, and
  Allen]{miller2003automatic}
Andrew~T Miller, Steffen Knoop, Henrik~I Christensen, and Peter~K Allen.
\newblock \href{http://ieeexplore.ieee.org/document/1241860/}{Automatic Grasp
  Planning Using Shape Primitives}.
\newblock In \emph{Proc.\ of the IEEE International Conference on Robotics and
  Automation (ICRA)}, pages 1824--1829, 2003.

\bibitem[Pinto and Gupta(2016)]{Pinto2016SupersizingHours}
Lerrel Pinto and Abhinav Gupta.
\newblock
  \href{http://ieeexplore.ieee.org/abstract/document/7487517/}{Supersizing
  self-supervision: Learning to grasp from 50k tries and 700 robot hours}.
\newblock In \emph{Proc.\ of the IEEE International Conference on Robotics and
  Automation (ICRA)}, pages 3406--3413, 2016.

\bibitem[Prattichizzo and Trinkle(2008)]{Prattichizzo2008Grasping}
Domenico Prattichizzo and Jeffrey~C. Trinkle.
\newblock
  \href{http://link.springer.com/10.1007/978-3-540-30301-5_29}{Grasping}.
\newblock In \emph{Springer Handbook of Robotics}, chapter~28, pages 671--700.
  Springer Berlin Heidelberg, 2008.

\bibitem[Redmon and Angelova(2015)]{Redmon2015Real-timeNetworks}
Joseph Redmon and Anelia Angelova.
\newblock \href{http://ieeexplore.ieee.org/document/7139361/}{Real-Time Grasp
  Detection Using Convolutional Neural Networks}.
\newblock In \emph{Proc.\ of the IEEE International Conference on Robotics and
  Automation (ICRA)}, pages 1316--1322, 2015.

\bibitem[Rubert et~al.(2017)Rubert, Kappler, Morales, Schaal, and
  Bohg]{Rubert2017OnSuccess}
Carlos Rubert, Daniel Kappler, Antonio Morales, Stefan Schaal, and Jeannette
  Bohg.
\newblock \href{http://ieeexplore.ieee.org/document/8202167/}{On the Relevance
  of Grasp Metrics for Predicting Grasp Success}.
\newblock In \emph{Proc.\ of the IEEE/RSJ International Conference of
  Intelligent Robots and Systems (IROS)}, pages 265--272, 2017.

\bibitem[Sahbani et~al.(2012)Sahbani, El-Khoury, and
  Bidaud]{sahbani2012overview}
Anis Sahbani, Sahar El-Khoury, and Philippe Bidaud.
\newblock
  \href{https://www.sciencedirect.com/science/article/pii/S0921889011001485}{An
  overview of 3D object grasp synthesis algorithms}.
\newblock \emph{Robotics and Autonomous Systems}, 60\penalty0 (3):\penalty0
  326--336, 2012.

\bibitem[Saxena et~al.(2008)Saxena, Driemeyer, and Ng]{Saxena2008RoboticVision}
Ashutosh Saxena, Justin Driemeyer, and Andrew~Y. Ng.
\newblock
  \href{http://journals.sagepub.com/doi/10.1177/0278364907087172}{Robotic
  Grasping of Novel Objects using Vision}.
\newblock \emph{The International Journal of Robotics Research (IJRR)},
  27\penalty0 (2):\penalty0 157--173, 2008.

\bibitem[Shimoga(1996)]{shimoga1996robot}
Karun~B Shimoga.
\newblock
  \href{http://journals.sagepub.com/doi/abs/10.1177/027836499601500302}{Robot
  Grasp Synthesis Algorithms: A Survey}.
\newblock \emph{The International Journal of Robotics Research (IJRR)},
  15\penalty0 (3):\penalty0 230--266, 1996.

\bibitem[Vahrenkamp et~al.(2008)Vahrenkamp, Wieland, Azad, Gonzalez, Asfour,
  and Dillmann]{Vahrenkamp2008VisualTasks}
N.~Vahrenkamp, S.~Wieland, P.~Azad, D.~Gonzalez, T.~Asfour, and R.~Dillmann.
\newblock \href{http://ieeexplore.ieee.org/document/4755985/}{Visual servoing
  for humanoid grasping and manipulation tasks}.
\newblock In \emph{Proc.\ of the International Conference on Humanoid Robots
  (Humanoids)}, pages 406--412, 2008.

\bibitem[Varley et~al.(2015)Varley, Weisz, Weiss, and
  Allen]{varley2015generating}
Jacob Varley, Jonathan Weisz, Jared Weiss, and Peter Allen.
\newblock \href{https://ieeexplore.ieee.org/document/7354004/}{Generating
  Multi-Fingered Robotic Grasps via Deep Learning}.
\newblock In \emph{Proc.\ of the IEEE/RSJ International Conference on
  Intelligent Robots and Systems (IROS)}, pages 4415--4420. IEEE, 2015.

\bibitem[Viereck et~al.(2017)Viereck, Pas, Saenko, and
  Platt]{Viereck2017LearningImages}
Ulrich Viereck, Andreas Pas, Kate Saenko, and Robert Platt.
\newblock \href{http://proceedings.mlr.press/v78/viereck17a.html}{Learning a
  visuomotor controller for real world robotic grasping using simulated depth
  images}.
\newblock In \emph{Proc.\ of the Conference on Robot Learning (CoRL)}, pages
  291--300, 2017.

\bibitem[Wang et~al.(2016)Wang, Li, Wang, and Liu]{Wang2016RobotNetworks}
Z.~Wang, Z.~Li, B.~Wang, and H.~Liu.
\newblock
  \href{http://ade.sagepub.com/lookup/doi/10.1177/1687814016668077}{Robot grasp
  detection using multimodal deep convolutional neural networks}.
\newblock \emph{Advances in Mechanical Engineering}, 8\penalty0 (9), 2016.

\bibitem[Yang et~al.(2016)Yang, Price, Cohen, Lee, and Yang]{yang2016object}
Jimei Yang, Brian Price, Scott Cohen, Honglak Lee, and Ming-Hsuan Yang.
\newblock \href{http://ieeexplore.ieee.org/document/7780397/}{Object Contour
  Detection with a Fully Convolutional Encoder-Decoder Network}.
\newblock In \emph{Proc.\ of the IEEE Conference on Computer Vision and Pattern
  Recognition (CVPR)}, pages 193--202, 2016.

\bibitem[{Yun Jiang} et~al.(2011){Yun Jiang}, Moseson, and
  Saxena]{Jiang2011EfficientRepresentation}
{Yun Jiang}, Stephen Moseson, and Ashutosh Saxena.
\newblock \href{http://ieeexplore.ieee.org/document/5980145/}{Efficient
  Grasping from RGBD Images: Learning using a new Rectangle Representation}.
\newblock In \emph{Proc.\ of the IEEE International Conference on Robotics and
  Automation (ICRA)}, pages 3304--3311, 2011.

\end{thebibliography}

\end{document}